 \RequirePackage[hyphens]{url}
 \documentclass[pmlr,twocolumn,10pt]{jmlr} 

\usepackage{booktabs}
\usepackage{siunitx}

\usepackage[switch]{lineno}

\newcommand{\equal}[1]{{\hypersetup{linkcolor=black}\thanks{#1}}}

\newsavebox\CBox

\theorembodyfont{\upshape}
\theoremheaderfont{\scshape}
\theorempostheader{:}
\theoremsep{\newline}

\jmlrvolume{LEAVE UNSET}
\jmlryear{2024}
\jmlrsubmitted{LEAVE UNSET}
\jmlrpublished{LEAVE UNSET}
\jmlrworkshop{Machine Learning for Health (ML4H) 2024} 

\title[Explainable Cell Graphs]{\vspace{-0.4cm}xCG: Explainable Cell Graphs for Survival Prediction in Non-Small Cell Lung Cancer}

\author{
\Name{Marvin Sextro}\equal{These authors contributed equally} \textsuperscript{\textnormal{1,2,3,4}} \Email{m.kleine.sextro@tu-berlin.de} \\
\Name{Gabriel Dernbach}\footnotemark[1] \textsuperscript{\textnormal{1,3,5}} \\
\Name{Kai Standvoss} \textsuperscript{\textnormal{1}} \\
\Name{Simon Schallenberg} \textsuperscript{\textnormal{5}} \\
\Name{Frederick Klauschen} \textsuperscript{\textnormal{3,5,6,7,8}} \\
\Name{Klaus-Robert Müller} \textsuperscript{\textnormal{2,3,9,10}} \\
\Name{Maximilian Alber} \textsuperscript{\textnormal{1,5}} \\
\Name{Lukas Ruff} \textsuperscript{\textnormal{1}} \\
\addr \textsuperscript{\textnormal{1}}Aignostics, Germany\\
\addr \textsuperscript{\textnormal{2}}Machine Learning Group, Technische Universität Berlin, Germany\\
\addr \textsuperscript{\textnormal{3}}BIFOLD -- Berlin Institute for the Foundations of Learning and Data, Germany\\
\addr \textsuperscript{\textnormal{4}}Konrad Zuse School of Excellence in Learning and Intelligent Systems (ELIZA), Germany\\
\addr \textsuperscript{\textnormal{5}}Institute of Pathology, Charité -- Universitätsmedizin Berlin, Germany\\
\addr \textsuperscript{\textnormal{6}}German Cancer Research Center (DKFZ) \& German Cancer Consortium (DKTK), Berlin \& Munich Partner Sites\\
\addr \textsuperscript{\textnormal{7}}Institute of Pathology, Ludwig-Maximilians-Universität München, Germany\\
\addr \textsuperscript{\textnormal{8}}Bavarian Cancer Research Center (BZKF), Germany\\
\addr \textsuperscript{\textnormal{9}}Max-Planck Institute for Informatics, Germany\\
\addr \textsuperscript{\textnormal{10}}Department of Artificial Intelligence, Korea University, Republic of Korea\\
}

\begin{document}

\maketitle

\begin{abstract}
Understanding how deep learning models predict oncology patient risk can provide critical insights into disease progression, support clinical decision-making, and pave the way for trustworthy and data-driven precision medicine. Building on recent advances in the spatial modeling of the tumor microenvironment using graph neural networks, we present an explainable cell graph (xCG) approach for survival prediction. We validate our model on a public cohort of imaging mass cytometry (IMC) data for 416 cases of lung adenocarcinoma. We explain survival predictions in terms of known phenotypes on the cell level by computing risk attributions over cell graphs, for which we propose an efficient grid-based layer-wise relevance propagation (LRP) method. Our ablation studies highlight the importance of incorporating the cancer stage and model ensembling to improve the quality of risk estimates. Our xCG method, together with the IMC data, is made publicly available to support further research.
\end{abstract}

\begin{figure}[t!]
\floatconts
  {fig:overview}
  {\caption{Overview of the xCG framework, including preprocessing steps, GNN for survival prediction, and GNN-LRP risk attribution.}}
  {\includegraphics[width=0.99\linewidth]{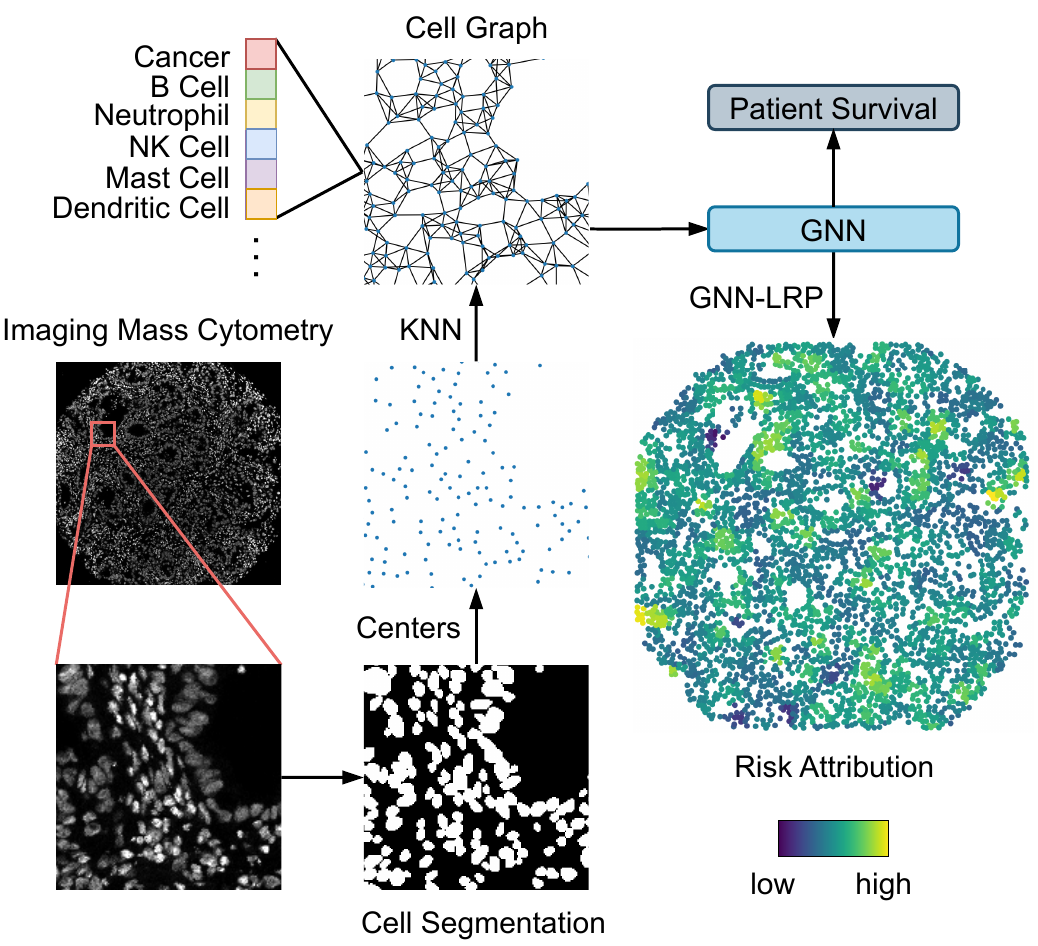}}
\end{figure}

\paragraph*{Data and Code Availability}\label{sec:data_code}
We publish our PyTorch implementation of xCG at \url{https://github.com/marvinsxtr/explainable-cell-graphs}. Our method is validated on a publicly available\footnote{\url{https://doi.org/10.5281/zenodo.7760826}} data cohort \citep{Sorin2023}.

\paragraph*{Institutional Review Board (IRB)}
This study was performed according to the ethical principles for medical research of the Declaration of Helsinki and approved by the Ethics Committee of the Charité University Medical Department in Berlin (EA4/243/21).

\section{Introduction}
Lung cancer remains the leading cause of cancer-related death, accounting for over 20\% of all cancer cases \citep{Siegel2021}. To improve patient outcomes, it is crucial to further advance our understanding of the disease mechanisms and identify more precise risk factors. 

Established risk factors such as the UICC8's TNM classification of malignant tumors \citep{brierley2017tnm} represent the standard of care but lack granularity for personalized treatment decisions, e.g.\ not covering immune system, molecular, or metabolic parameters, often resulting in unnecessary side effects and rendering treatment insufficient.

Advances in spatially resolved single-cell technologies now allow us to explore the tumor microenvironment (TME) in unprecedented detail \citep{Sorin2023}. Leveraging these technologies, graph neural networks (GNNs) have shown promise in modeling the TME in several cancer types, including lung cancer \citep{Zhou2019, Wang2022, Nakhli2023, Zhang2024}. However, while recent studies have begun to incorporate explainability into graph-based models \citep{Sureka2020, Jaume2020, Hu2024, Zhang2024}, explaining risk factors remains a significant challenge in the medical field due to the large scale of relevant graphs.

In this paper, we present the following key contributions to address this challenge:
\begin{itemize}
    \item \textbf{Modality-Agnostic Survival Prediction}: We propose a versatile GNN framework that can (a) handle multiple tissue samples and graphs per patient, (b) incorporate multiple cell-level feature domains such as marker expression, tumor region-segmentation and patient-level clinical metadata, and (c) is capable of survival regression and classification. Our implementation in PyTorch is publicly available.
    \item \textbf{Scalable XAI for Cell Graphs}: We introduce a novel efficient grid-based GNN-LRP method for cell graphs that enables high-resolution risk attribution at the cell level.
    \item \textbf{Enhanced Risk Assessment}: Our ablation studies show that combining cancer stage fusion and model ensembling significantly improves the accuracy and reliability of risk assessments.
\end{itemize}

\section{Methods}

\subsection{Data and Preprocessing}

\paragraph{Data} We use a dataset published in \citet{Sorin2023} consisting of single tissue spots of $1.0\ \text{mm}^2$, obtained from 416 patients with adenocarcinoma of the lung. Spots are stained with a 35-plex imaging mass cytometry (IMC) panel, from which 17 distinct cell phenotypes were derived.

The cohort includes clinical metadata like overall survival and the UICC8 cancer stage, categorizing patients into early (I-II) and late (III-IV) stages. For survival classification, we adapt the two categories proposed by \cite{Sorin2023}: short-term ($\leq 36$ months) and long-term ($> 36$ months) survival. Patients with a survival time of less than 36 months but without registered death events were excluded, as they could have died in either period.

\paragraph{Preprocessing} As shown in \figureref{fig:overview}, the centers of mass of the cell segmentation masks are used to determine cell positions. For each tissue sample, we construct a cell graph where each cell is represented as a node. Each node is characterized by a one-hot encoded vector representing the cell phenotype. Biologically resembling mutual interactions between proximal cells, edges are established by $k$-nearest neighbors (KNN) fit with $k=3$.

\subsection{Models and Training}
\label{sec:models_training}

\paragraph{Survival Regression}
Our proposed cell graph encoder architecture (\figureref{fig:survival_regression_architecture}) builds on the sparse hierarchical graph classifier framework presented in \citet{Cangea2018}.
To enable this architecture to input multiple tissue samples per patient, we incorporate attention-based MIL pooling as described in \citet{Ilse2018}, so the input becomes a set of cell graphs, $\mathcal{X} = \{\mathcal{G}_1, ..., \mathcal{G}_K\}$. We made this generalization of the architecture since the use case of multiple spots per patient is often given, as was the case for a proprietary study cohort of ours.

In the first stage, the model computes an embedding for each graph individually  $\mathcal{H} = \{\vec{h}_1, ..., \vec{h}_K\}$ by alternating between Graph Isomorphism Network (GIN) layers \citep{Xu2018} and top-k pooling layers \citep{Cangea2018}.
To integrate clinical metadata for enhanced risk assessment, we fuse the UICC8 cancer stage to the cell graph representations by addition.
In the second stage, to produce a patient-level representation $\vec{h}_\mathcal{X}$, attention-based MIL-pooling is applied, which allows the model to prioritize the most relevant graphs when creating the overall patient representation.
Finally, the aggregated patient representation is passed through a feed-forward network that is responsible for predicting survival risk. 

\paragraph{Survival Classification}
To further simplify the explainability of the model (see \sectionref{sec:explainability}), we also reformulate the problem as a binary survival classification task and distinguish between short-term ($\leq 36$ months) and long-term ($> 36$ months) survival.

To implement this classification task, we use a 3-layer graph isomorphism network (GIN) without graph pooling, adapted from the implementation by \cite{Schnake2022}. Instead of using attention-based multiple instance learning (MIL), we average the GIN logits across the multiple cell graphs per patient to simplify relevance propagation. Future work may consider implementing xMIL-LRP \citep{hense2024xmil}. Additionally, we omit the cancer stage fusion for our explainability analysis, as its relevance would be non-localized and not contribute to the spatial explanation.

\paragraph{Training}
We train both our survival regression and classification models using stratified nested cross-validation over five folds, training for 50 epochs. Hyperparameter optimization is performed on inner validation folds, with the learning rate chosen from $\gamma \in \{5\text{e}{-5}, 1\text{e}{-5}, 5\text{e}{-6}\}$. To improve the robustness of our risk predictions, particularly towards variability caused by different model initialization, we ensemble the risk predictions of five separate survival regression models. Each model is trained with a different random seed, and we compute the final prediction by taking the mean of the outputs.

\subsection{Explainability}
\label{sec:explainability}

\begin{table*}[t]
\floatconts
  {tab:uicc_ablation}%
  {\caption{Ablation of UICC8 cancer stage fusion and model ensembling for our survival regression model. We report the C-index and its standard deviation over five seeds, averaged over five test folds.}}%
  {\begin{tabular}{c c c c}
   \toprule
   \textbf{UICC8 Fusion} & \textbf{UICC8 Baseline} & \textbf{GIN} & \textbf{Ensemble} \\
   \midrule
   \checkmark & 0.568 $\pm$ 0.005 & 0.559 $\pm$ 0.019 & 0.593 $\pm$ 0.033 \\
   - & - & 0.507 $\pm$ 0.022 & 0.518 $\pm$ 0.043 \\
   \bottomrule
   \end{tabular}}
\end{table*}

Ranking-based survival models, such as Cox regression, are widely used in medical research to predict patient outcomes \citep{jozwiak2024cox}. However, interpreting how these rank-based models make predictions is difficult, due to their implicit handling of survival times.

To illustrate, layer-wise relevance propagation (LRP) propagates the predicted risk score backwards through the network, assigning relevance (attributions) to input features. One challenge is that these attributions can be both positive and negative, depending on how we define the 'zero point' in the model's target range. As \citet{Letzgus2023} discusses, the point we set as the baseline (or zero point) determines whether a feature is seen as increasing or decreasing the predicted risk.

Furthermore, the magnitude of these attributions changes with the scale of the target range. Without a clear definition of both the baseline and the scale, it becomes difficult to interpret the model's explanations. This is particularly problematic for ranking-based models, as they are trained using an implicit ranking loss that does not allow us to set an explicit reference point and scale.
Therefore, we only generate explanations for the survival classification model discussed in \sectionref{sec:models_training}, where these reference points are made explicit. For more details on interpreting model explantions see \sectionref{sec:interpreting_explanations}

We use sGNN-LRP \citep{Xiong2022}, an optimized variant of GNN-LRP \citep{Schnake2022} tailored to subgraph attribution. This method reduces the computational complexity from exponential $\mathcal{O}(|\mathcal{S}|^L)$ to linear $\mathcal{O}(L|\mathcal{S}|^2)$, where $L$ is the number of layers in the network and the number of nodes in the subgraph. While the graphs considered in \citet{Xiong2022} only span at most hundreds of nodes, our cell graphs can reach tens of thousands of nodes. We significantly reduce the memory requirements by exploiting the sparse connectivity of the KNN-based graph adjacency matrix and rewrite sGNN-LRP with sparse matrix multiplications using PyTorch Sparse\footnote{\url{https://github.com/rusty1s/pytorch_sparse}}.

To further reduce memory and compute costs of high-resolution risk attributions, we split the global attribution task into local subtasks utilizing a shifted-grid approximation approach. The cell graph is partitioned into a square grid and the subgraph relevance is calculated for each tile and normalized by the number of cells, mitigating over-weighting of densely populated areas. The grid is then repeatedly shifted in the $x$- and $y$-directions by a stride $s$, such that the average over strides results in a smooth heatmap of cell-level risk attributions.

\begin{figure*}
\floatconts
  {fig:risk_attribution_phenotypes}
  {\caption{Median LRP relevance over cases, localized by cell phenotype for (\textit{a}) short ($\leq 36$ months) and (\textit{b}) long ($> 36$ months) survival. Phenotypes are sorted by their median LRP relevance. We perform a permutation test over 1,000 iterations (*** $p < 0.001$, ** $p < 0.01$, * $p < 0.05$).}}
  {%
    \subfigure[Short survival.][c]{
      \includegraphics[width=0.48\linewidth]{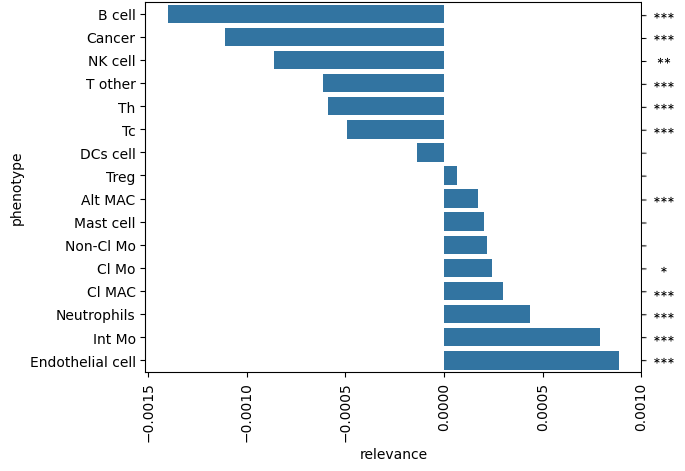}
    }
    \subfigure[Long survival.][c]{
      \includegraphics[width=0.48\linewidth]{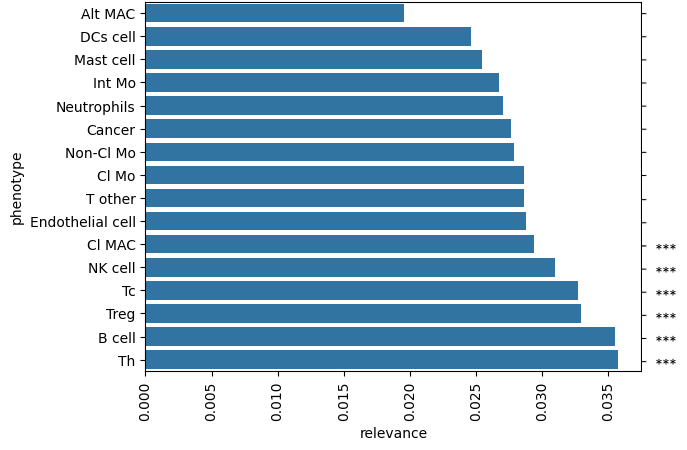}
    }
  }
\end{figure*}

\section{Results and Discussion}

\subsection{Survival Regression}
We evaluate the performance of our cell graph survival regression model by reporting the C-index \citep{harrell1982}, a metric of a model's ability to correctly rank survival times, for three different scenarios: the unmodified standard UICC8 clinical baseline, our GNN-based survival regression model, and the mean risk ensemble of our model, each averaged over five test folds.
The results presented in \tableref{tab:uicc_ablation} show an improvement in C-index from 0.568 to 0.593 when using our GNN ensemble, highlighting the model's ability to capture complex interactions within the TME and effectively advancing the clinical baseline.
The improved performance of the ensemble over the single GIN model shows that model ensembling can mitigate the variance found in individual model predictions and produce more robust, generalizing survival estimates.

Removing cancer-stage fusion leads to a significant decrease in C-index. This result highlights that the integration of clinical staging information remains a crucial context for accurate survival prediction.

\subsection{Survival Classification}

We train a survival classification model to generate risk attribution heatmaps for cell graphs following the methodology described in \sectionref{sec:explainability}. Our model was trained on five different seeds and evaluated across five test folds, resulting in a binary AUROC of 0.700 $\pm$ 0.028 without cancer stage fusion.

Plotting the risk attributions computed via our grid-based GNN-LRP approach on a per-cell basis, we achieve high-resolution heatmaps as depicted in \figureref{fig:risk_attribution_heatmaps}. We visualize min-max normalized relevance and compare risk attribution heatmaps between spots with short and long survival. We observe strongly localized regions of high (yellow) and low (blue) risk attribution, which can be traced back to known cell phenotypes.

\paragraph{Tying Risk Attributions to Disease Biology}\label{sec:risk_quantitative_eval}
We calculate the median relevance attributed to the cells of a phenotype for long and short survivors separately and perform a permutation test over 1,000 iterations. The comparison of the median LRP relevance across cases (\figureref{fig:risk_attribution_phenotypes}) according to cell phenotype reveals a notable difference between long and short survivors. Among the long survivors, several immune cell phenotypes such as T helper cells, B cells, regulatory T cells, or cytotoxic T cells are particularly relevant ($p < 0.001$), aligning with the findings of previous studies in the literature \citep{debatin2024prognostic, denkert2018tumour, galon2006type, laumont2022tumour, wieland2021defining, hu2021atlas}. In contrast, the particular relevance of neutrophils in short-term survivors ($p < 0.001$) is consistent with the established association between increased neutrophil counts and poorer prognosis in NSCLC patients \citep{ilie2012predictive}.

\section{Conclusion}

In this work, we presented a framework for explaining large-scale cell graphs using high-resolution cell-level risk attributions. We further showed that cancer stage fusion and model ensembling can improve survival prediction based on cell graphs. Moreover, we see an indication that cell graphs are capable of learning spatial TME features that are beyond the classical understanding of what is relevant for prognosis in the clinic.

\acks{We would like to thank Thomas Schnake and Ping Xiong for the helpful discussions and the reviewers for their constructive feedback, which helped to improve this paper.

This work was in part supported by the German Ministry for Education and Research (BMBF) under Grants 01IS14013A-E, 01GQ1115, 01GQ0850, 01IS18025A, 031L0207D, and 01IS18037A. M.S.\ acknowledges support by the Konrad Zuse School of Excellence in Learning and Intelligent Systems (ELIZA). K.R.M.\ was partly supported by the Institute of Information \& Communications Technology Planning \& Evaluation (IITP) grants funded by the Korea government (MSIT) (No.\ 2019-0-00079, Artificial Intelligence Graduate School Program, Korea University and No.\ 2022-0-00984, Development of Artificial Intelligence Technology for Personalized Plug-and-Play Explanation and Verification of Explanation).}

\bibliography{jmlr-sample}

\appendix
\clearpage

\section{Methods}\label{appendix:methods}

\subsection{Model and Training}

We use the AdamW optimizer \citep{Loshchilov2017} with the default parameters $\beta_1=0.9$, $\beta_2=0.999$ and $\epsilon=1\text{e}{-8}$ to train our models. \tableref{tab:hyperparameters} lists other hyperparameters used for training our survival regression and classification models. During training, the learning rate is reduced by a cosine annealing schedule. Our training setup is implemented using \texttt{hydra-zen} \citep{Soklaski2022} to be easily configurable and reproducible. We use PyTorch Geometric \citep{Fey2019} for the implementation of our GNN-based survival regression model.

Our survival regression models are trained using the Cox negative partial log-likelihood loss \citep{Kvamme2019}:
\[
    \mathcal{L}(\vec{\theta}) = \frac{1}{N_{\delta = 1}} \sum_{i\,:\,\delta_i = 1} \log \left( \sum_{j \in \mathcal{R}_i} \exp \left[\hat{h}_{\vec{\theta}}(\vec{x}_j) - \hat{h}_{\vec{\theta}}(\vec{x}_i)\right] \right)
\]

All models were trained on an NVIDIA L4 GPU with 24 GB of memory.

\begin{table}
\floatconts
  {tab:hyperparameters}%
  {\caption{Hyperparameters used for survival regression and classification. Due to implementation constraints, we use online learning for survival classification.}}%
  {\begin{tabular}{l c c}
   \toprule
   \textbf{Hyperparameter} & \textbf{Value(s)} \\
   \midrule
   Learning rate & $\{5\text{e}{-5}, 1\text{e}{-5}, 5\text{e}{-6}\}$ \\
   Batch size & 16 (1) \\
   Hidden dimension & 64 \\
   Message passing layers & 3 \\
   Number of epochs & 50 \\
   \bottomrule
   \end{tabular}}
\end{table}

\begin{figure}
\floatconts
  {fig:survival_regression_architecture}
  {\caption{Illustration of our proposed bag of cell graphs GNN architecture for survival regression. The cell graphs are constructed from the KNN neighborhood of the individual cell of patients. The UICC8 cancer stage is fused before the MIL pooling. MIL pooling is either implemented with attention-based pooling \citep{Ilse2018} for better performance or mean pooling, for simpler interpretation.}}
  {\includegraphics[width=0.4\linewidth]{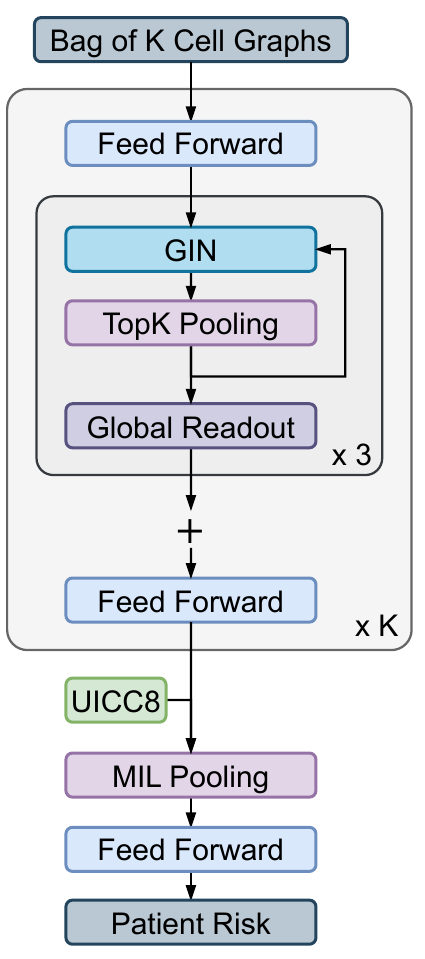}}
\end{figure}

\section{Evaluation}

\subsection{Scalability}

\begin{figure}
\floatconts
  {fig:runtime}
  {\caption{Comparison of wall-clock runtime between the naive GNN-LRP implementation and our grid-based sGNN-LRP method for synthetically generated cell graphs with different numbers of nodes. The runtimes are averaged over three repetitions, with the vertical lines indicating the standard deviation.}}
  {\includegraphics[width=\linewidth]{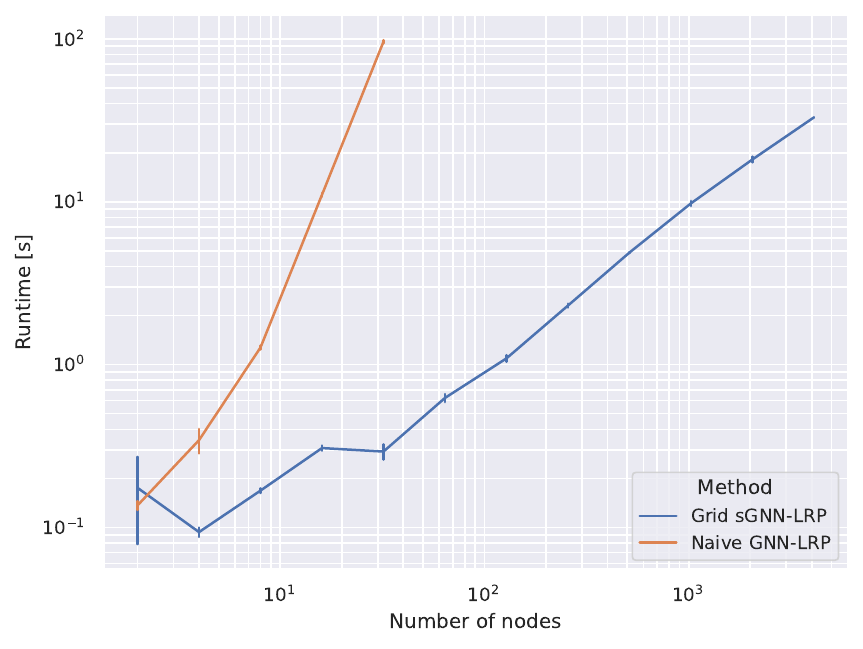}}
\end{figure}

\paragraph{Runtime Comparison}
To evaluate the scalability of our grid-based sGNN-LRP method to large cell graphs, we compare our approach to a naive GNN-LRP implementation \citep{Schnake2022} computing relevances exhaustively. For this purpose, we synthetically generate cell graphs by sampling a given number of nodes uniformly in the unit circle and applying KNN as before. \figureref{fig:runtime} shows wall-clock runtimes averaged over three repetitions for increasing numbers of nodes. While the naive implementation is prohibitively slow, reaching a runtime of over one minute at 32 nodes, our grid-based method is capable of explaining graphs with more than a thousand nodes in seconds.

\paragraph{Memory Usage}
Storing the full adjacency matrix as a dense matrix results in a memory requirement of $\mathcal{O}(n^2)$, where $n$ is the number of nodes. Due to the particular sparsity of our graphs, albeit with high numbers of nodes, we especially profit from the reduced memory requirements of the PyTorch Sparse implementation, which does not store the zero entries of the adjacency matrix.

\subsection{Interpreting Model Explanations}
\label{sec:interpreting_explanations}
The layer-wise relevance propagation (LRP) input attribution heatmaps can be interpreted similarly to those produced by the input-times-gradient rule, as both methods reflect the sensitivity of the model to input perturbations. In our case, the input is a one-hot vector that can only activate or deactivate the attribution without affecting its sign, allowing an independent discussion of the role of the gradient.

To interpret the sign, consider a scenario where two output neurons represent the softmax probabilities for positive and negative predictions. A positive gradient in the neuron for the positive prediction indicates evidence for a positive outcome (e.g.\ long survival). Conversely, a positive gradient in the negative prediction neuron signals evidence for a negative outcome (e.g.\ short survival).

Therefore, the interpretation of positive attributions depends on the context of the specific classification result. In summary, these heatmaps should be seen as the evidence that the model uses to justify its prediction.

\subsection{Qualitative Evaluation of Risk Attributions}

\begin{figure*}
\floatconts
  {fig:risk_attribution_roi}
  {\caption{Regions of interest showing the spatial distribution of cell phenotypes next to the corresponding risk attribution heatmaps in (\textit{a}) long and (\textit{b}) short survival. High values of attribution indicate positive evidence for the respective model decision, for more details on interpreting explanation heatmaps see \sectionref{sec:interpreting_explanations}.}}
  {%
    \subfigure[Long survival.][c]{
      \includegraphics[width=\linewidth]{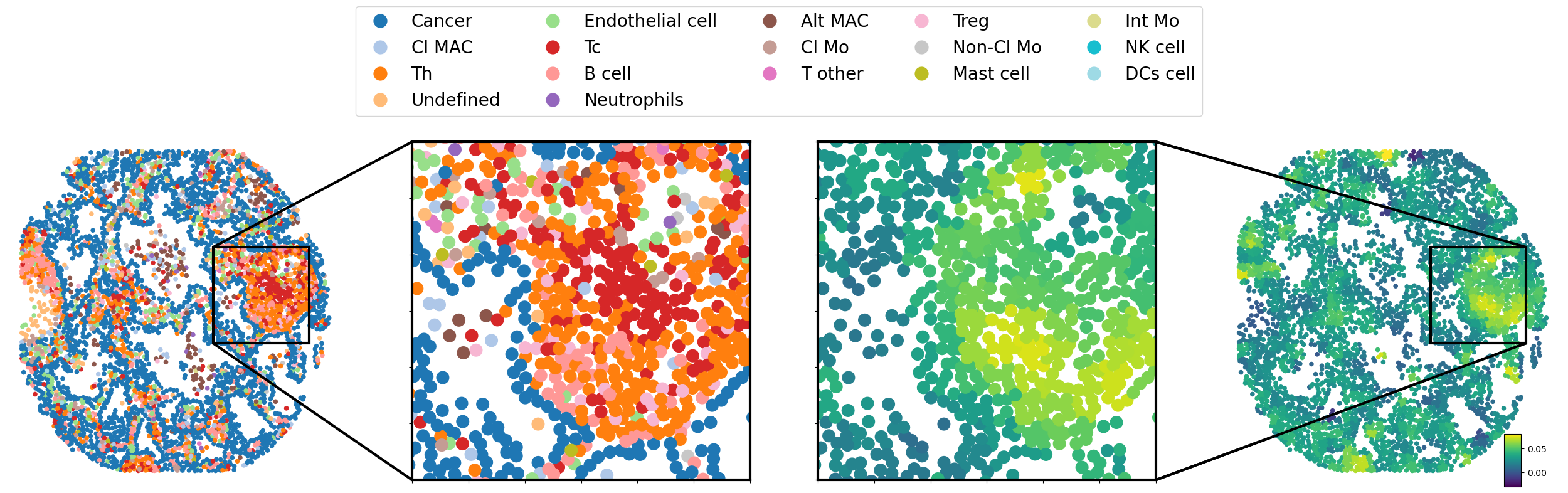}
    }
    \subfigure[Short survival.][c]{
      \includegraphics[width=\linewidth]{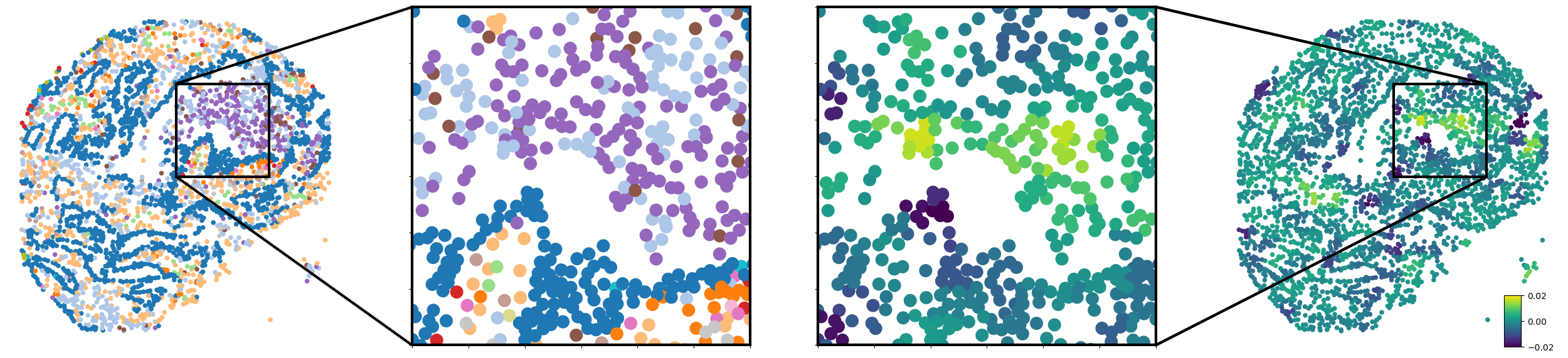}
    }
  }
\end{figure*}

In addition to quantitatively evaluating risk attributions among phenotypes across cases described in \sectionref{sec:risk_quantitative_eval}, we perform a qualitative analysis by visualizing cell types and risk attribution heatmaps for selected spots. \figureref{fig:risk_attribution_roi} shows zoomed-in regions of interest for exemplary spots. For long survival, we see that high relevance is predominantly assigned to immune cells (T helper cells, B cells, regulatory T cells, and cytotoxic T cells), while the surrounding cancer tissue is assigned lower relevance. Looking at short survival, we see that high relevance is attributed to an area enriched in neutrophils and classical macrophages, while the surrounding cancer tissue is of lower relevance. These findings are in line with our quantitative analysis across cases in \sectionref{sec:risk_quantitative_eval}, showing our method can give meaningful insight into the TME for individual tissue spots, which is consistent with existing domain knowledge.

\begin{figure*}
\floatconts
  {fig:risk_attribution_heatmaps}
  {\caption{Exemplary risk attribution heatmaps generated by our grid-based GNN-LRP method for (\textit{a}) short ($\leq 36$ months) and (\textit{b}) long ($> 36$ months) survival. Attributions are computed with a tile size $t=0.05$ mm and a stride $s=0.025$ mm. To ensure comparability within survival classes, we min-max normalize risk attributions in the intervals $[-0.02, 0.02]$ and $[-0.03, 0.08]$ for short and long survival, respectively. High values of attribution indicate positive evidence for the respective model decision, for more details on interpreting explanation heatmaps, see \sectionref{sec:interpreting_explanations}. For a side-by-side comparison to the annotated phenotypes of highly attributed cells, see \figureref{fig:risk_attribution_roi}}}
  {%
    \subfigure[Short survival.][c]{
      \includegraphics[width=0.45\linewidth]{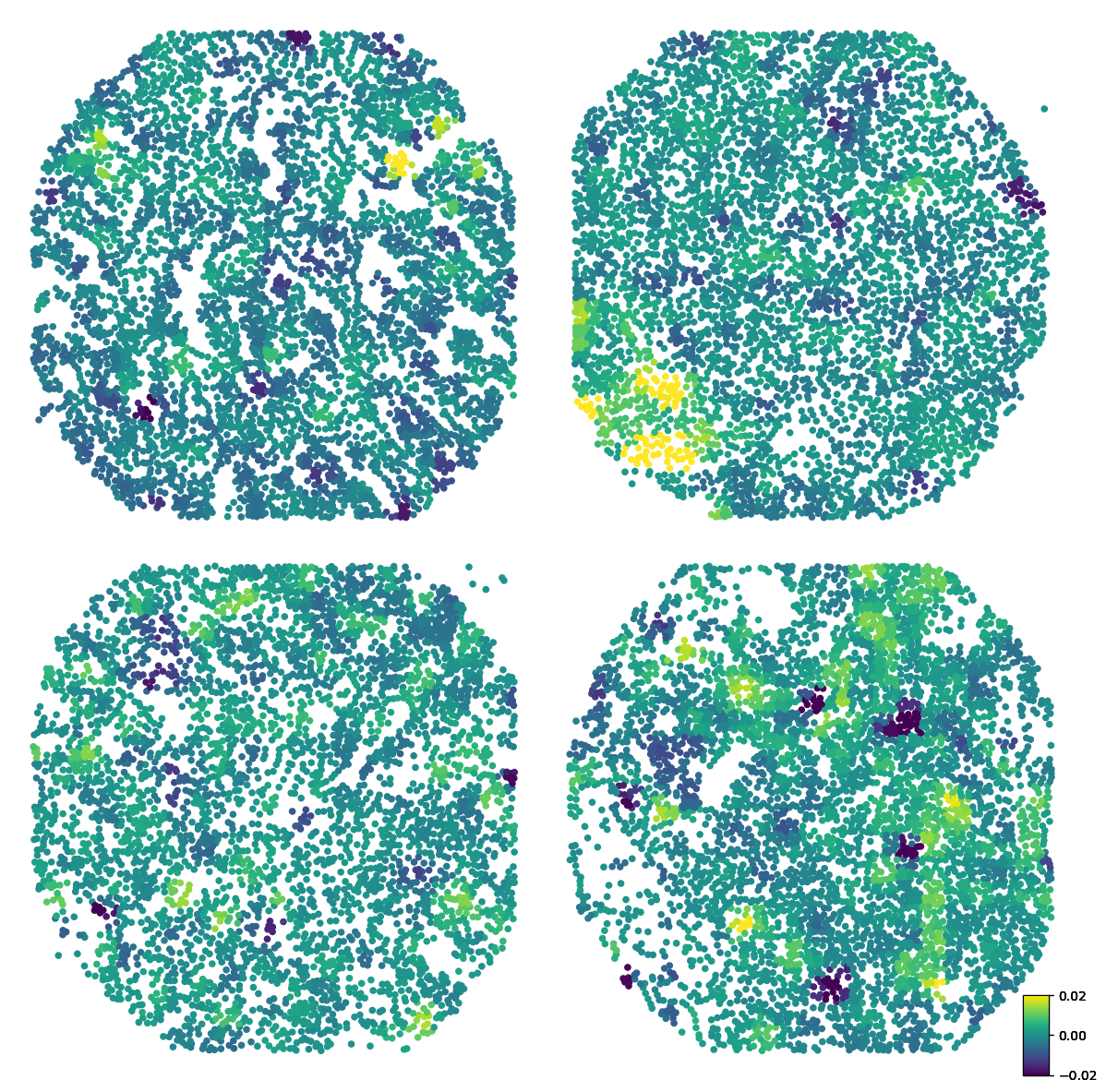}
    }
    \subfigure[Long survival.][c]{
      \includegraphics[width=0.45\linewidth]{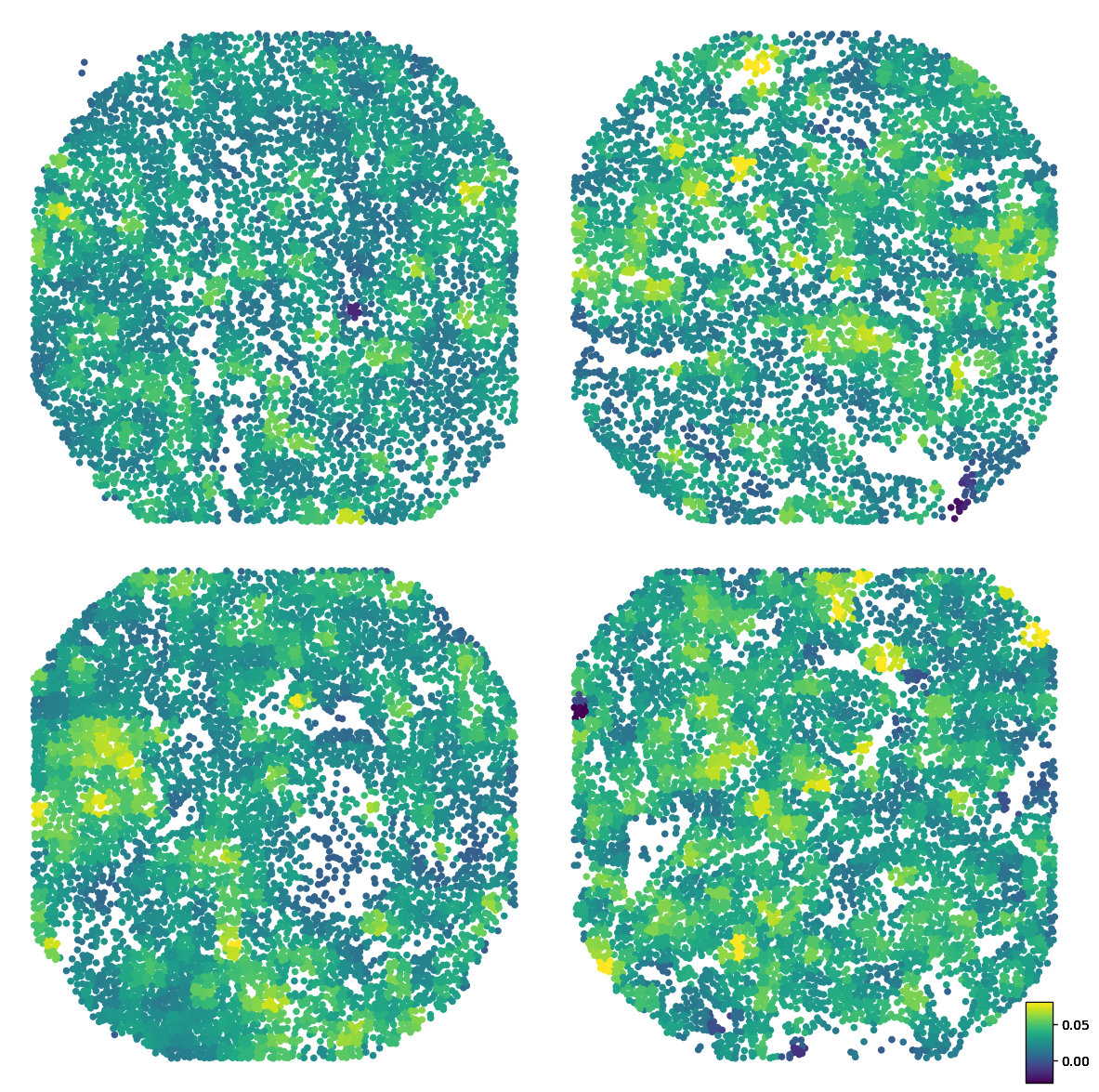}
    }
  }
\end{figure*}

\end{document}